\documentclass[a4paper,11pt,twocolumn,twoside]{article}
\usepackage{graphicx}
\usepackage{sepln_en}
\usepackage{fullname}
\usepackage[utf8]{inputenc}
\usepackage[english]{babel}
\usepackage{amsmath}
\usepackage{amsfonts}
\usepackage{xspace}
\usepackage{hyperref}

\input epsf
\makeatletter
\renewcommand{\@biblabel}[1]{} % elimina [n]
\makeatother
\newcommand{\ie}{\textit{i}.\textit{e}., }
\newcommand{\eg}{\textit{e}.\textit{g}., }
\newcommand{\alg}{\textit{ALC}\xspace}

\title{Fine-grained auxiliary learning for real-world product recommendation}
\author {\textbf{Mario Almagro,} \textbf{Diego Ortego,} \textbf{David Jiménez} \\
NielsenIQ\\
\{mario.almagro, diego.ortego, david.jimenez\}@nielseniq.com\\
}

\seplnkey{retrieval, coverage, regularization, auxiliary.}

\seplnabstract{Product recommendation is the task of recovering the closest items to a given query within a large product corpora. 
Generally, one can determine if top-ranked products are related to the query by applying a similarity threshold; exceeding it deems the product relevant, otherwise manual revision is required.
Despite being a well-known problem, the integration of these models in real-world systems is often overlooked. In particular, production systems have strong coverage requirements, \ie a high proportion of recommendations must be automated.
In this paper we propose \alg, an \textbf{A}uxiliary \textbf{L}earning strategy that boosts \textbf{C}overage through learning fine-grained embeddings. Concretely, we introduce two training objectives that leverage the hardest negatives in the batch to build discriminative training signals between positives and negatives. We validate \alg using three extreme multi-label classification approaches in two product recommendation datasets; LF-AmazonTitles-131K and Tech\&Durables (proprietary), demonstrating state-of-the-art coverage rates when combined with a recent threshold-consistent margin loss.}

\makeatletter
\providecommand{\@seplntranstitle}{}      % internal macro used by sepln_en.sty
\providecommand{\seplntranstitle}[1]{\gdef\@seplntranstitle{#1}} % (optional) user macro
\makeatother

\firstpageno{1}

\begin{document}

% la siguiente instrucción sólo se debe usar si el abstract sobrescribe el texto
% la longitud variará según se necesite

\setlength\titlebox{20cm} % se aumenta el tamaño del espacio reservado para datos de título

\label{firstpage} \maketitle

%\begin{abstract}
%Resumen del artículo con una sangría a izquierda y derecha de 0.32
%cm, justificado por ambos lados, con tamaño de fuente 11.
%
%\end{abstract}

\section{Introduction}
% Traditional recommender systems and limitations when it comes to real production systems and automation rates.
Information retrieval is the task of recovering the most relevant documents in response to a query~\cite{Hambarde23}. These systems are normally composed of two sequential stages. First, a light-weight retrieval model recovers a limited number of documents (from a large database) that are close to a given query, and then a re-ranking stage exploits fine-grained dependencies between query-document pairs to refine the recommendation. The former is specially designed for efficiency and the objective is to ensure that all positive documents appear within the top-k retrieved items. The latter, is normally a more capable but computationally intensive model that decides whether the presented query-document pair is positive or not. Therefore, the retriever narrows down the scope of the candidates for the second stage, reducing the overall computation and latency.

% Recommender systems are normally composed of two sequential stages: 1) a light-weight retrieval model that is capable of recovering the top-k closest documents to a query from a large document database, and 2) a binary classification model in charge of learning fine-grained dependencies between query-document pairs. The former is specially designed for efficiency and the objective is to ensure that all positive documents appear within the top-k items. The latter, is normally a more capable but computational intensive model in charge of taking a binary decision on whether the presented query-document pair is positive or not. We normally use the retriever in a way to reduce the scope of the candidates for the second stage, thus reducing the overall computation and latency.

Traditional retrieval models~\cite{Karpukhin20} are trained to learn an embedding representation space suitable to measure similarities between queries and documents. These models focus on delivering high-quality recommendations by providing an ordered list of items based on their similarity scores with the query. However, real-world problems require that the recommended list is actually relevant~\cite{ZhangQ2024}, which is often challenging due to several factors. For instance, the scarcity of information in the query text, the absence of relevant items in the list or simply the low quality of the models can lead to retrieving irrelevant items. 
Providing retrieval or recommendation models with the capability to reject a set of hypotheses is of great interest in real-world systems such as Retrieval Augmented Generation (RAG) pipelines~\cite{ZhangRAFT2024,Wei25}, where adding noisy context to a Large Language Model (LLM) can make the generated output to fail dramatically. Moreover, in decision making processes, we need to ensure that the recommendations satisfy high quality requirements or, otherwise, rely on human intervention to take the decision.

% Traditional retrieval models are trained to learn a metric space suitable to measure similarities between query and document embeddings. These models focus on delivering high quality recommendations, providing an ordered list of documents based on their similarities to the query. However, there are several factors that influence the relevance of the recommendations. For instance, due to the lack of capacity of existing models, the scarcity of information in the query text or missing documents on the database, one can find many situations where the closest items are not related to the query. % the that does not guarantee that the closest item is related to the query. 
% Providing retrieval models with the capability to reject a set of hypothesis is of great interest in real-world recommender systems. In recent Retrieval Augmented Generation (RAG) pipelines adding noisy context to a Large Language Model (LLM) can make the generated output fail dramatically, while in decision making processes we need to ensure that the recommendations have a minimum level of quality or, otherwise, rely on human intervention to take the decision.

% Why coverage is relavant for real-world systems.
In the context of open-world retrieval systems, the concept of coverage~\cite{Franc23} becomes crucial, \ie the percentage of accepted samples for a given performance requirement based on the predicted score. Usually, acceptance is determined when exceeding a certain similarity score threshold between query and candidate document. This threshold is set to ensure that automatically generated predictions satisfy a performance requirement. Thus, high coverage rates reduce the amount of non-trusted predictions that need human revision. However, setting a threshold to reject candidate items is challenging due to the so-called threshold inconsistency problem~\cite{ZhangQ2024}, \ie different semantic concepts might have different distributions and, therefore, require specific thresholds to identify relevant items.

With the exponential growth of the different e-commerce channels, product recommendation~\cite{Luo24,Dahiya25} is receiving increased attention from the research community. Product information retrieval systems are typically characterized by extremely large product spaces, short and often incomplete descriptions, and a high level of granularity. Moreover, models trained on general-purpose data often struggle with the specialized vocabulary, resulting in limited performance.
For instance, this type of challenges can be observed in the following example where an input query such as "Oreo pb crm choc sdnwch 17oz" needs to be mapped to semantically relevant products like "Chips ahoy orig choc chip 13oz" and "Lotus bisc sndwch crm 7.76oz" from a set of millions of candidate products. Additionally, descriptions with missing information such as “Coca-Cola Zero”, where details like volume or packaging are not available, can result in low similarity scores, significantly harming coverage.

In this context, extreme multi-label classification (XMC) methods strike a balance between scalability and efficiency~\cite{Chang20}, mainly using encoder-only Transformer architectures~\cite{Gupta24,Dahiya25}. These approaches are normally based on two building blocks: a backbone encoder trained with metric learning and, extreme classifiers trained using one-vs-all strategies. XMC models have demonstrated state-of-the-art performance in existing benchmarks for product recommendation, being capable of retrieving similar items or labels quite effectively.
However, the separability of similarity scores, and consequently the coverage, is often overlooked, hindering the applicability to real-world systems.
In this paper, we propose \alg, a novel regularization technique based on auxiliary learning that uses two pretext objectives during training designed to improve the separability of similarity scores between queries and in-batch items that are sampled using hard-negative sampling techniques.
The resulting embeddings are capable of capturing fine-grained details between queries and the hardest in-batch negative labels, promoting hard negatives to be smoothly distributed in the embedding space. 
We empirically demonstrate that when \alg is applied in combination with the TCM (Threshold-Consistent Margin) loss~\cite{ZhangQ2024}, both regularization terms complement each other and lead to remarkable improvements in coverage results while having little negative impact on precision for product recommendation.

Our main contributions are as follows:
\begin{itemize}
    \item We propose \alg a novel auxiliary learning strategy that improves the separability of relevant and irrelevant items. 
    \item We successfully validate the applicability of the TCM regularization, originally tested in image retrieval scenarios, to text-based product recommendation.
    \item Our results demonstrate that \alg and TCM regularization together can obtain state-of-the-art results for the coverage of XMC methods, while mostly keeping precision performance.
\end{itemize}

%By adding regularization strategies that enforce a clear distinction between positive and negative pairs in the learned representation space, we aim to enhance the consistency of similarity scores.
%Imposing constraints on the learned metric forces the model to better separate positive matches from negatives, ultimately increasing the coverage of high-confidence product matches in the retrieval process. 
%Our experiments in two product recommendation datasets show that 
%Our approach demonstrates that forcing such consistency through proper regularization can lead to more accurate and efficient product matching, without relying on a subsequent classifier stage.

\section{Related work}

% Brief intro to recommender systems and introduce coverage challenge
Search tasks are fundamental in NLP as demonstrated by the variety of related problems addressed in the literature, \eg information retrieval~\cite{Hambarde23}, recommender systems~\cite{Raza24}, product recommendation~\cite{Dahiya25} or product matching~\cite{Peeters22}. Information retrieval~\cite{Hambarde23,Ma24} aims at ranking the most relevant documents in response to a query, playing an important role in web search, question answering, fact verification and retrieval-augmented generation pipelines. Differently, recommender systems~\cite{Raza24,Liu25} are algorithms designed to suggest items (books, movies, products, or content) to users based on their preferences. These systems typically exploit user and item profiles during the recommendation process. In product search or recommendation~\cite{Luo24,Dahiya25}, queries are short and typically built from keywords that are mostly combinations of product attribute information. Additionally, the search space is exclusively built from products (structured data containing attribute information such as brand, color or size), as opposed to web search where the search space might include web pages, videos or other types of unstructured items. Finally, product matching~\cite{Peeters22,Almagro22} is a particular case of product recommendation, where the related items to find in a product corpus are the exact same products provided as queries. In this matching task, query and candidate items usually come from different data sources, \eg retailers.

Despite the variety of search tasks, deep learning has unified the strategy to address them by using metric learning based approaches \cite{Schroff15,Kim20,Zeng22,Peeters22,Ma24,Dahiya25}. The purpose of metric learning in computer vision and NLP~\cite{Kim20,Zhang21} is to learn a score that measures the similarity between a pair of image or text samples. In particular, these methods work by pulling together representations of the same concept and pushing apart representations for different ones. Therefore, the final output of the search is a ranked list of documents or items based on their similarity score. Furthermore, in real-world production systems, the similarity between item pairs is thresholded so that those exceeding the threshold will be automated as relevant~\cite{ZhangQ2024}, while the rest will go through manual revision pipelines.
Thresholding is challenging due to the threshold inconsistency problem~\cite{ZhangQ2024}, \ie different semantic concepts might have different distributions, thus requiring varying similarity thresholds to achieve top performance.

The threshold inconsistency can be addressed using post-hoc calibration methods~\cite{Pakdaman15,Guo17,Naghiaei22,Kweon24}, which adjust the similarity scores after training. However, these methods require prior knowledge about the test distribution and can be inefficient and lack robustness. In contrast, reducing threshold inconsistencies during training is an approach that does not assume any test distribution and focuses on producing calibrated similarity scores that better reflect relevancy. In particular, there are image retrieval methods of this kind~\cite{Veit20,LiuJ22,ZhangQ2024}. 
The authors in~\cite{Veit20} propose a cross-example softmax loss that encourages all queries to be closer to their matching images than to all non-matching ones. This loss leads to a more calibrated similarity score that better reflects a measure of relevancy.
Liu et. al~\cite{LiuJ22} introduce a Threshold-consistency penalty in the loss. In particular, they propose to divide the images of each mini-batch into several domains, whose specific calibration thresholds are estimated using in-batch negative pairs and later exploited to adaptively adjust the loss weights for each domain.
More recently, the authors in~\cite{ZhangQ2024} propose the Threshold-Consistent Margin (TCM) loss, a regularization defined over hard pairs by penalizing query-to-relevant and query-to-irrelevant image pairs whose similarity score is low or high, respectively, promoting uniformity in representation structures across classes.

In this work, we focus on product recommendation using encoder-based extreme multi-label classification (XMC)~\cite{Gupta24,Dahiya25}, where state-of-the-art methods ignore the calibration of their recommendations. In particular, we propose two novel auxiliary learning tasks to improve coverage results. Additionally, we bring the TCM regularization~\cite{ZhangQ2024} from image retrieval and achieve further improvements.

% % Methods addressing automation/coverate (TCM)
% Metric learning-based retrieval systems, such as those using InfoNCE and Triplet Margin losses, have shown promise in improving retrieval accuracy. However, challenges related to coverage by thresholding similarity scores remain, particularly when dealing with ambiguous or noisy data. Various regularization strategies have been proposed to address these issues, such as..., but a clear solution to improve the consistency and separability of learned similarity scores in product matching tasks has not yet been fully explored.

% % why OURS is different from previous works (we move TCM-tbc)
% Our work is based on Siamese architectures that learn robust text representations and label prototypes based on contrastive learning. We add two novel regularization techniques—Threshold-Consistent Margin (TCM) and Classifier-driven top-K re-Ranking (ClassiRank)—to improve the separability of products and refine the retrieval coverage.

% TBD

\section{Proposed approach \label{sec:method}} 

We propose \alg, a novel regularization technique based on auxiliary learning designed to improve the embedding-based similarity score distribution. This regularization is suitable for encoder-only Transformer architectures where both queries and items are processed independently through the backbone (bi-encoder configuration). 
We hypothesize that learning a more discriminative metric space for hard negatives improves inter-class separability and, consequently the ability to distinguish relevant from irrelevant similarity scores between embeddings.
In particular, we use \alg in extreme multi-label classification for product recommendation, where a Transformer encoder is trained using deep metric learning and in-batch hard negative mining. These approaches normally rely on contrastive learning objectives (\eg InfoNCE or triplet loss) and sample the top labels associated to semantically similar queries in a batch. Note that, we force one positive query-label pair to appear within the top items, thus reinforcing more uniform intra-class compactness needed to improve calibration (we refer the reader to~\cite{ZhangQ2024} for more details about the impact of inter/intra-class separability in calibration).
%We propose \alg, a receipt of three loss terms that regularize the product recommendation training to produce similarity scores that better separate relevant and irrelevant products.

\subsection{Setup}
Consider~$\mathcal{D} = \{q_i, \mathcal{P}_i\}_{i=1}^{Q}$ be a training set with a set of $Q$ queries $\mathcal{Q}=\{q_i\}_{i=1}^{Q}$, where each query $q_i$ has a positive label set $\mathcal{P}_i$. For any $q_i$, we define the space of $L$ labels $\mathcal{Y}=\{l_r\}_{r=1}^{L}$ as a super-set of the positive $\mathcal{P}_i = \{p_j\}_{j=1}^{P_i}$ and negative $\mathcal{N}_i = \{n_k\}_{k=1}^{N_i}$ sets. $P_i$ and $N_i$ are the number of positive and negative labels associated to $q_i$, respectively. For positive label sampling, we uniformly select one positive label for each query in the batch. On the other hand, for negative mining we use the method proposed in \cite{Dahiya23}, which builds query batches $\mathcal{B}_t = \{q_b, \; b=1, ..., B\}$ of size $B$ at each training step $t$ by selecting semantically related queries. Thus, semi-hard negative labels naturally arise for every query $q_i$ using the positive labels associated to every other query in the batch, \ie $n_i = {p_j, \: \forall j \neq i \in \mathcal{B}_t}$. We also experiment with the negative mining strategy proposed at \cite{Xiong20}, that directly computes the pool of hardest negatives for each query. Then, we uniformly sample one negative from each query pool. 

% Consider~$\mathcal{D} = \{q_i, \mathcal{P}_i\}_{i=1}^{Q}$ be a training set with a set of $Q$ queries $\mathcal{Q}=\{q_i\}_{i=1}^{Q}$, where each query $q_i$ has a positive label set $\mathcal{P}_i$ (to improve clarity on the notation, we refer to document and labels indistinctly). We define the space of $L$ label documents $\mathcal{Y}=\{l_r\}_{r=1}^{L}$ for any $q_i$ to be a super-set of the positive $\mathcal{P}_i = \{p_j\}_{j=1}^{P_i}$ and the negative sets $\mathcal{N}_i = \{n_k\}_{k=1}^{N_i}$. $P_i$ and $N_i$ are the number of positive and negative labels associated to $q_i$, respectively. For each training step $t$ during training, we build batches of size $B$ as $\mathcal{B}_t = \{q_b, \; b=1, ..., B\}$ using semantically related queries. Thus, hard negatives for every query $q_i$ can be identified using the positive labels associated to every other query in the batch, \ie $n_i = {p_j, \: \forall j \neq i \in \mathcal{B}_t}$. 

We pose XMC as a maximum inner product search task between query and label embeddings to accurately predict the positive labels for every query $q_i$. To accomplish this, we aim to learn a function $f_{\theta}: (\mathcal{Q}, \; \mathcal{Y})  \rightarrow \mathbb{R}^{d}$, where $\theta$ denotes the parameters of a neural network model that encodes the given textual representation into a $d$-dimensional sentence embedding. This function is used to encode every query $q_i$ and label $l_r$ into $\mathbf{h}_{q_i}$ and $\mathbf{h}_{l_r}$. Note that $\mathbf{h}_{l}$ can be defined as $\mathbf{h}_{p}$ and $\mathbf{h}_{n}$ to distinguish between positive and negative text-based label embeddings, respectively.

\subsection{Auxiliary learning for fine-grained embeddings}

\begin{figure}[t]
    \centering
    \includegraphics[width=0.95\linewidth]{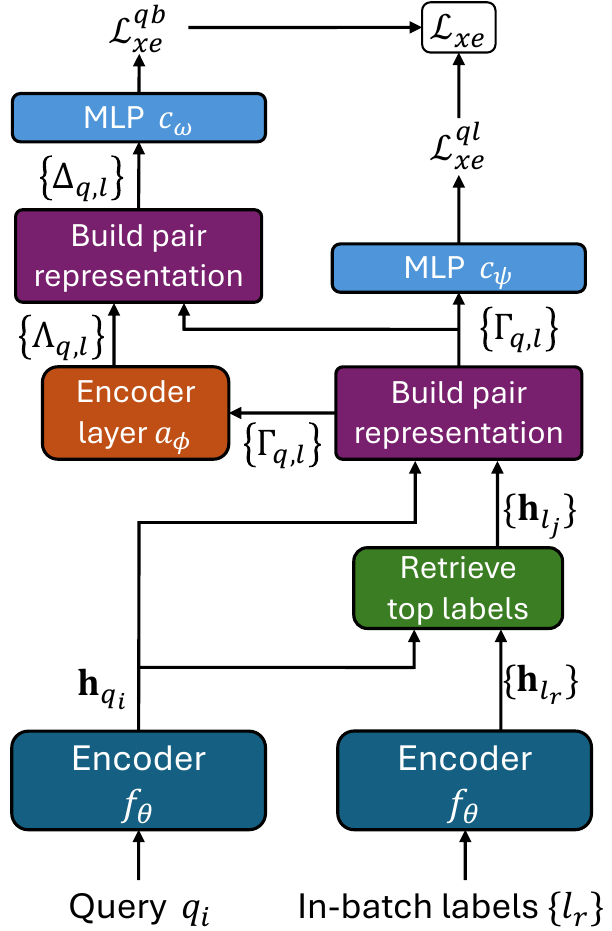}
    \caption{General diagram of \alg.}
    \label{fig:general_diagram}
\end{figure}

\alg is composed of two loss functions that are jointly optimized:
\begin{equation}
    \begin{split}
        \mathcal{L}_{xe} = \beta_1\mathcal{L}_{xe}^{q\,l} + \beta_2\mathcal{L}_{xe}^{q\,b},
    \end{split}
\end{equation}
where $\beta_1$ and $\beta_2$ control the weight of the regularizations. On the one hand, we optimize query and label embeddings to be able to discriminate positives from the hardest negatives in the batch ($\mathcal{L}_{xe}^{q\,l}$). On the other hand, we create a batch-aware representation of queries and labels ($\mathcal{L}_{xe}^{q\,b}$) and train the model on the same discrimination task. The second term helps the model discover new patterns by contextualizing the query-label representations with a blocking of labels, \ie the $K-1$ hardest in-batch negatives and a positive pair. \alg promotes queries and the negative labels in the blocking to be pushed apart while pulling the positive in the blocking close to the query embedding. 
In Figure~\ref{fig:general_diagram}, we present the general diagram of our approach.

We build pair representations following the same methodology of~\cite{Reimers19}, \ie we concatenate each independent embedding, their difference and their element-wise multiplication to build query-label pairs representations $\Gamma(\mathbf{h}_{q}, \mathbf{h}_{l})$ as follows:
\begin{equation}
    \Gamma(\mathbf{h}_{q}, \mathbf{h}_{l}) = \{\mathbf{h}_{q}, \mathbf{h}_{l}, |\mathbf{h}_{q}-\mathbf{h}_{l}|, \mathbf{h}_{q}\odot \mathbf{h}_{l}\}
\end{equation}
In the remaining of this section, we relax the notation of $\Gamma(\mathbf{h}_{q}, \mathbf{h}_{l})$ and refer to it as $\Gamma_{q,l}$. We subsequently feed the query-label representation to an MLP $c_{\psi}$ that acts as a binary classifier to determine the positive/negative class for each pair. Then, our first auxiliary loss term in \alg regularizes XMC training by optimizing the $\psi$ parameters of this classifier as follows:
\begin{equation}
    \begin{split}
        \mathcal{L}_{xe}^{q\,l} = -\frac{1}{QK} \sum_{i=1}^Q \sum_{j=1}^K (y_{q_i,l_j}\log(p^{\Gamma}_{q_i,l_j}) \\
        + (1 - y_{q_i,l_j})\log(1 - p^{\Gamma}_{q_i,l_j})).
    \end{split}
\end{equation}
The auxiliary loss $\mathcal{L}_{xe}^{q\,l}$ uses a binary cross-entropy objective, where $K$ is the number of elements in the blocking for each query $q_i$, $p^{\Gamma}_{q_i,l_j}$ is the sigmoid-normalized logit obtained after passing the representation $\Gamma_{q_i,l_j}$ through the MLP classifier $c_{\psi}$ and $y_{q_i,l_j}$ is the ground-truth class value of 0 and 1 for positive and negative query-label pairs, respectively. 
% A binary cross-entropy optimization is used to correlate pair representations $S_{q,k}$ with matches $y_{q,k}$.

% \begin{equation}
%     p^c_{q,k} = sigmoid(S_{q,k} W_C)
% \end{equation}
% The \alg regularization uses classifier-driven predictions to refine the text representation of the top-k candidates retrieved by the model. This approach maximizes the probability of predicting correct matches based on the overlaps features items.

In contrast, we define the second auxiliary loss term $\mathcal{L}_{xe}^{q\,b}$ to exploit cross-pair interactions within the $K$ in-batch pairs selected for the blocking of each query. In particular, we introduce a Transformer encoder block $a_{\phi}: \mathbb{R}^{K\times d^{\Gamma}}  \rightarrow \mathbb{R}^{K\times d^{\Gamma}}$ to contextualize each query-label pair with its corresponding $K$ pairs.
Batch-aware pair representations $\Lambda_{q,l}$ are built for each query-label pair by forwarding $d^{\Gamma}$-dimensional representation $\Gamma_{q,l}$ through $a_{\phi}$. We then build the representations $$\Delta_{q,l}=\{\Gamma_{q,l},\ \Lambda_{q,l},\ |\Gamma_{q,l}-\Lambda_{q,l}|,\ \Gamma_{q,l}\odot \Lambda_{q,l}\}$$ aiming at exploiting the differences of query-label pairs before and after their contextualization.
As in the first auxiliary term, we feed $\Lambda_{q,l}$ pair representations to the binary classifier $c_{\omega}$, which is an MLP with parameters $\omega$ that determines the positive/negative class for each pair. Then, the second auxiliary loss term of \alg is defined as:
\begin{equation}
    \begin{split}
        \mathcal{L}_{xe}^{q\,b} = -\frac{1}{QK} \sum_{i=1}^Q \sum_{j=1}^K (y_{q_i,l_j}\log(p^{\Delta}_{q_i,l_j}) \\
        + (1 - y_{q_i,l_j})\log(1 - p^{\Delta}_{q_i,l_j})).
    \end{split}
\end{equation}
Again, this auxiliary loss function $\mathcal{L}_{xe}^{q\,b}$ uses a binary cross-entropy objective, where $p^{\Delta}_{q_i,l_j}$ is the sigmoid-normalized logit obtained after passing through the $c_{\omega}$ MLP the representation $\Delta_{q_i,l_j}$.
We add \alg objective $\mathcal{L}_{xe}$ to the overall XMC methods optimization and further equip XMC with the TCM regularization in any loss term where metric learning is conducted. Concretely, we use as base methods NGAME~\cite{Dahiya23}, DEXA~\cite{Dahiya23b}) and PRIME~\cite{Dahiya25}. Overall, the optimization can be summarized as:
\begin{equation}
    \mathcal{L} = \mathcal{L}_{base} + \mathcal{L}_{xe},
\end{equation}
where $\mathcal{L}_{base}$ refers to a generic XMC method loss function.

Both the TCM and \alg regularization methods are designed to improve separability during metric learning. TCM enforces positive and negative similarities to lie within certain ranges (see Subsection \ref{subsec: TCM}), while \alg promotes an embedding space for hard negatives with improved inter-class separability that enables a better thresholding of similarity scores. Together, both regularization techniques improve the precision-coverage trade-off, thus producing models that are better prepared for real-world product recommendation systems.

\subsection{Threshold-Consistent Margin \label{subsec: TCM}}
The Threshold-Consistent Margin (TCM) loss function proposed at~\cite{ZhangQ2024} is designed to penalize situations where query-to-relevant and query-to-irrelevant image
pairs have, respectively, a low and high similarity score. 
In particular, this regularization is imposed on hard positives and hard negatives, \ie those pairs that do not respect certain similarity ranges.

In this work, we bring this idea from image retrieval and apply it to product recommendation, where we find it to be very useful as well. The TCM regularization function can be defined as follows:
\begin{equation}
    \begin{split}
        \mathcal{L}_{TCM} = \frac{1}{|\mathcal{S}^+|} \sum_{s\in\mathcal{S}^+} (m^{+}-s) \\
        + \frac{1}{|\mathcal{S}^-|} \sum_{s\in\mathcal{S}^-} (s-m^{-}),
    \end{split}
\end{equation}
where $s=\mathbf{h}_{q}\cdot\mathbf{h}_{l}$ denotes a query-label similarity score and
$\mathcal{S}^+$ and $\mathcal{S}^-$ are, respectively, the set of hard positives and hard negatives whose cardinality is denoted as $|\cdot|$. The hard positive set $\mathcal{S}^+$ contains any in-batch query-positive label pair whose similarity is below the positive margin $m^{+}$, while the hard negative set $\mathcal{S}^-$ pairs contains any in-batch query-negative label pair whose similarity is below the negative margin $m^{-}$.
Therefore, this regularization adds the average difference from the similarity score to the corresponding margin as a penalty, thus increasing the overall loss value accordingly.

When used, we add the TCM loss to any contrastive loss term of the adopted XMC methods and name it as $\mathcal{L}_{base}^{TCM}$. For instance, adding the TCM loss to PRIME implies including it in the three triplet loss terms of the method.

% To achieve this, we add a new loss term that increases when the similarity between a query and a positive sample falls below a threshold (denoted as minimum margin) or when the similarity between a query and a negative sample exceeds a threshold (denoted as maximum margin). In doing so, TCM encourages the model to ensure that the similarities between positive pairs are sufficiently high, while the similarities between negative pairs are kept low. This penalty forces the model to adjust its embeddings such that the positive and negative samples are more clearly separated, improving the overall separability within the learned metric space.

\section{Experimental work}
\subsection{Datasets and metrics}

\begin{table}[t]
\begin{center}
\begin{tabular}{lll}
\hline
{\bf Statistic} & {\bf AM} & {\bf TD}\\ \hline
Accessibility & Public & Private\\
Label size & 131,073 & 215,745\\
Train size & 294,805 & 2,218,194\\
Test size & 134,835 & 118,886\\
Avg. query/label & 5.15 & 10.96\\
Avg. label/query & 2.29 & 1.01\\ \hline
\end{tabular}
\end{center}
\caption{\label{tabla_data_statistics}Dataset statistics for the product recommendation data of LF-AmazonTitles-131K (AM) and Tech\&Durables (TD).}
\end{table}

We benchmark our method on two product recommendation datasets: LF-AmazonTitles-131K (publicly available\footnote{\url{http://manikvarma.org/downloads/XC/XMLRepository.html}}) and Tech\&Durables (proprietary). Both datasets contain short-text product descriptions for queries and labels. 
The overview of these datasets is provided in Table~\ref{tabla_data_statistics}.

The high granularity of certain products makes manual annotation challenging, significantly increasing the cost of human annotators for these tasks. As a result, coverage becomes especially important.
The evaluation metrics reported are Precision and Coverage for the top-1 recommendation. With these metrics we asses both accuracy and the relevancy of the top-1 recommendations. In particular, coverage is calculated using target P@1 performance of 85\% for LF-AmazonTitles-131K dataset and 95\% for Tech\&Durables. Three random seeds have been used during training to evaluate the datasets.

\subsection{Implementation details}
We use the state-of-the-art XMC method PRIME~\cite{Dahiya25} and the earlier methods DEXA~\cite{Dahiya23b} and NGAME~\cite{Dahiya23} as our base models. Unless otherwise stated, we adopt their default hyperparameters and configurations. These methods use a backbone encoder to learn similarities between queries and labels, thus being suitable to measure synergies with our proposal.
In particular, we selected the pre-trained all-MiniLM-L6-v2 as our backbone model~\cite{Reimers19}. Note that both MLPs in \alg use the same architecture: one linear layer followed by layer normalization, dropout equals to 0.1, GeLU activation function and a final linear layer.

We train for 300 epochs in LF-AmazonTitles-131K with a batch size of 2048 and learning rate equals to $2\,e{-4}$, while in Tech\&Durables we train for 150 epochs with a batch size of 3200 and a learning rate of $3\,e^{-4}$. The maximum sequence length is set to 32 tokens in all experiments.

For the TCM regularization, we set the positive margin $m^{+}$ and the negative margin $m^{-}$ to the recommended values in~\cite{ZhangQ2024} (0.8 and 0.5, respectively). As introduced in Section~\ref{sec:method}, we sum the TCM regularization to each of the triplet loss terms of the methods to enhance the separability of similarity scores learned (we refer the reader to the original articles for further details~\cite{Dahiya23,Dahiya23b,Dahiya25}). 
We use two different negative sampling techniques depending on the dataset. For LF-AmazonTitles-131K, we use the sampler proposed at~\cite{Dahiya23}, while for Tech\&Durables data we adopt ANCE sampler~\cite{Xiong20}.
As for regularization weights, we use $\beta_1$=1.0 and $\beta_2$=0.5, reducing the impact of blocking information. Blocking size $K$ was set to 5. These values were chosen on the basis of preliminary experiments, ensuring an appropriate trade-off between precision and coverage.

\begin{table}[t]
\begin{center}
\begin{tabular}{lcc}
\hline
{\bf Method} & {\bf P@1} & {\bf C@1} \\ \hline
\multicolumn{3}{c}{\bf LF-AmazonTitles-131K (public)} \\ \hline
\small{NGAME ($\mathcal{L}_{\text{base}}$)} & 40.40 & 0.0\\
\small{DEXA ($\mathcal{L}_{\text{base}}$)} & \underline{41.25} & 0.02\\
\small{PRIME ($\mathcal{L}_{\text{base}}$)}  & \textbf{43.26} & 1.77\\ \hline
\small{NGAME ($\mathcal{L}_{\text{base}}^{\text{TCM}} \ +\ \mathcal{L}_{xe}$)} & 39.08 & 0.0\\
\small{DEXA ($\mathcal{L}_{\text{base}}^{\text{TCM}} \ +\ \mathcal{L}_{xe}$)} & 41.22 & \underline{13.64}\\
\small{PRIME ($\mathcal{L}_{\text{base}}^{\text{TCM}} \ +\ \mathcal{L}_{xe}$)} & 40.26 & \textbf{14.59}\\ \hline
\multicolumn{3}{c}{\bf Tech.\&Durables (private)}\\ \hline
\small{NGAME ($\mathcal{L}_{\text{base}}$)} & 77.69 & 12.20\\
\small{DEXA ($\mathcal{L}_{\text{base}}$)} & 78.94 & 32.47\\
\small{PRIME ($\mathcal{L}_{\text{base}}$)}  & \textbf{80.83} & 46.74\\ \hline
\small{NGAME ($\mathcal{L}_{\text{base}}^{\text{TCM}} \ +\ \mathcal{L}_{xe}$)} & 77.79 & 58.81\\
\small{DEXA ($\mathcal{L}_{\text{base}}^{\text{TCM}} \ +\ \mathcal{L}_{xe}$)} & 78.79 & \underline{62.92}\\
\small{PRIME ($\mathcal{L}_{\text{base}}^{\text{TCM}} \ +\ \mathcal{L}_{xe}$)} & \underline{80.05} & \textbf{65.08}\\ \hline
\end{tabular}
\end{center}
\caption{\label{tab:XMC}Performance of XMC methods and effect of combining them with \alg and TCM regularizations. Precision (P@1) and Coverage (C@1) values demonstrate that, despite advances in P@1, C@1 can sometimes be dramatically low for vanilla XMC methods. Adding both regularizations greatly improves C@1, while having little impact on P@1. Coverage is computed based on target P@1 of 85\% for LF-AmazonTitles-131K and 95\% for Tech\&Durables.}
\end{table}

\subsection{Results}

\begin{figure}[t]
    \centering
    \includegraphics[width=0.98\linewidth]{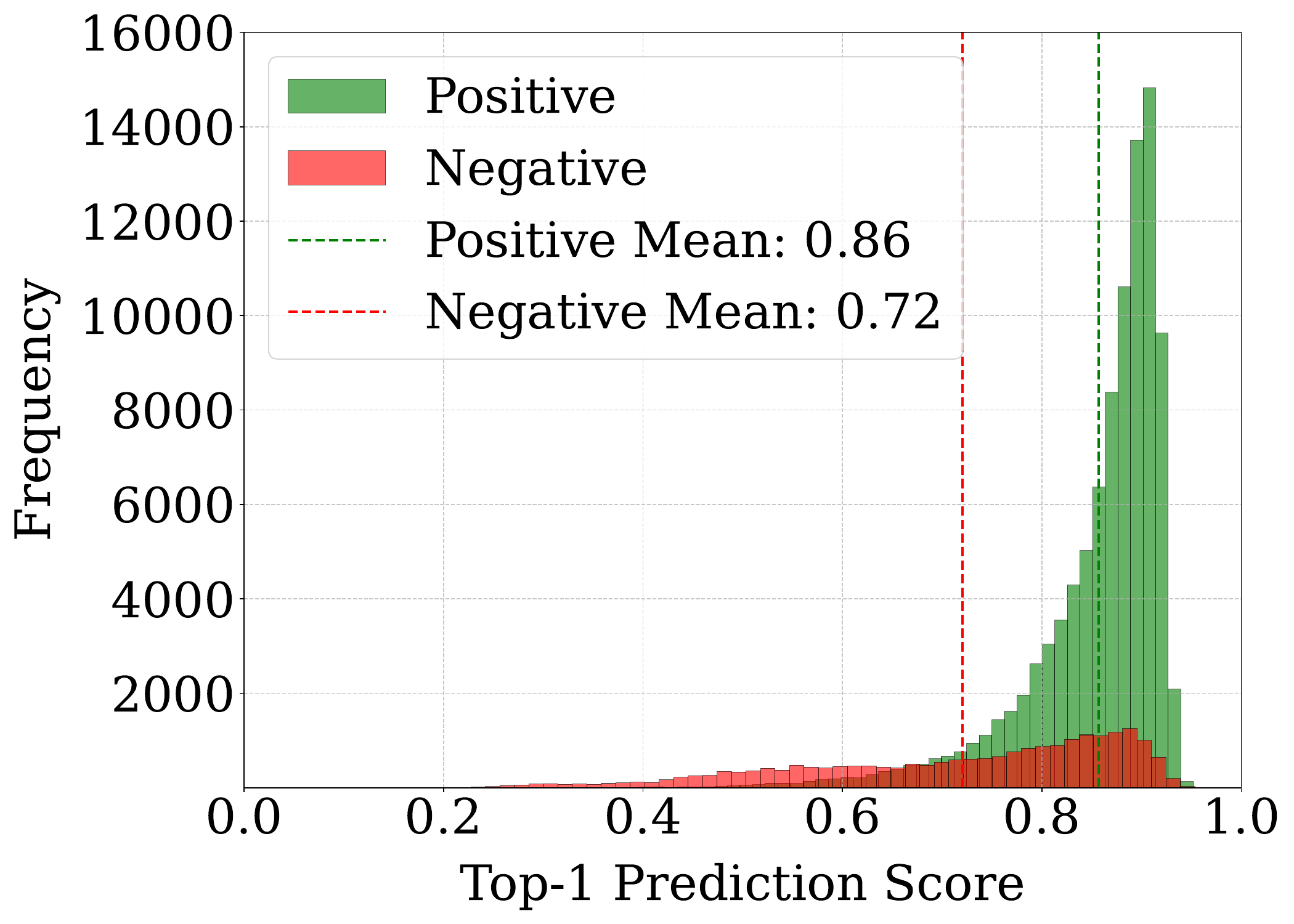}
    \includegraphics[width=0.98\linewidth]{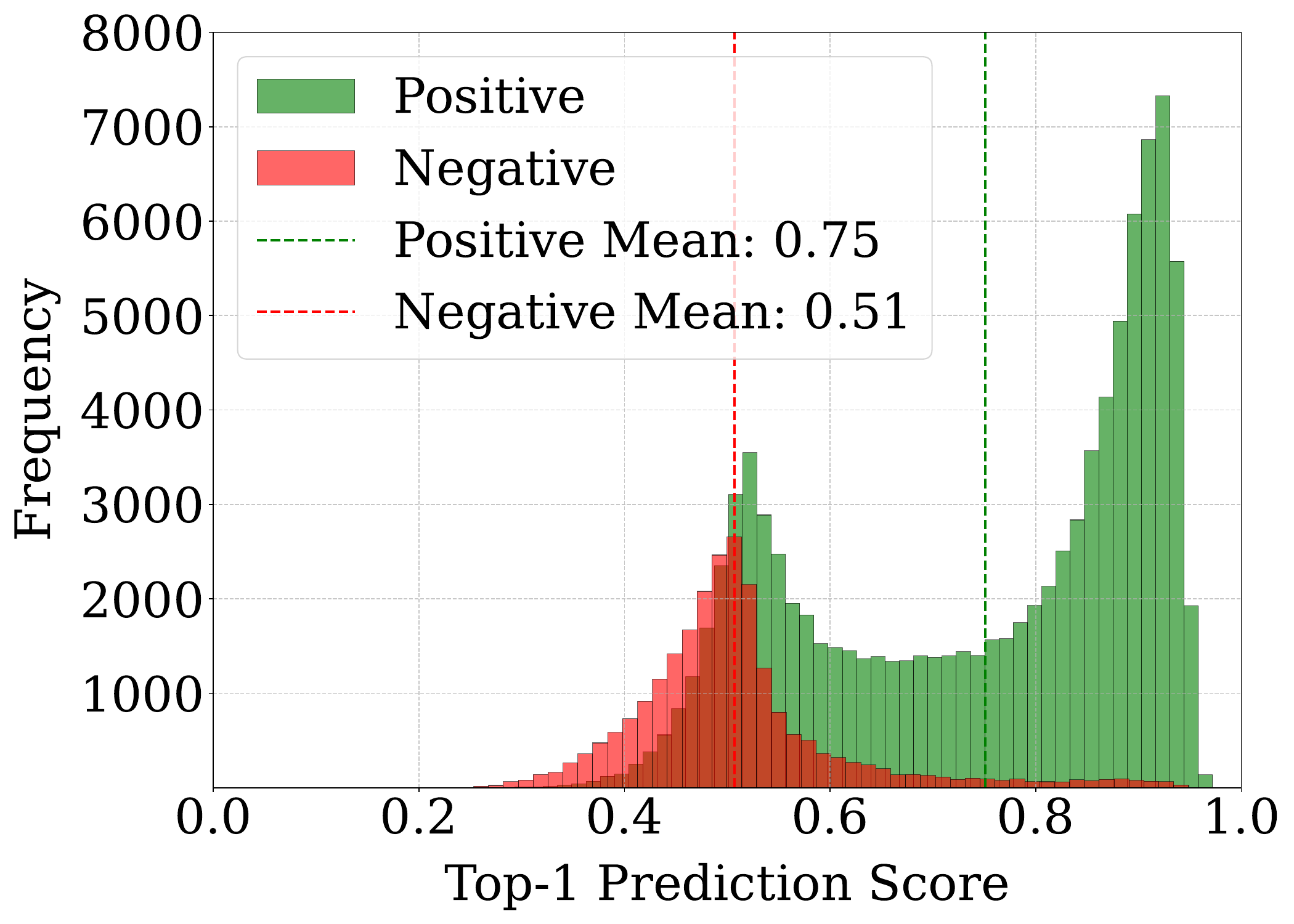}
    \caption{Histogram of top-1 predicted scores for PRIME (top) and PRIME together with \alg and TCM regularizations (bottom) in Tech\&Durables dataset.}
    \label{fig:distributions}
\end{figure}

We seek improving the coverage of XMC methods for product recommendation. In Table~\ref{tab:XMC} we show that coverage rates for existing XMC methods (NGAME, DEXA and PRIME) using the corresponding base objective functions $\mathcal{L}_{\text{base}}$ is indeed very low.
The proposed strategy using two auxiliary learning tasks in combination with the TCM regularization~\cite{ZhangQ2024} ($\mathcal{L}_{\text{base}}^{\text{TCM}} + \mathcal{L}_{\text{xe}}$) greatly improves coverage with little drops in precision. The only exception where improvements are not seen is for NGAME in LF-AmazonTitles-131K, where 85\% precision is not achieved and, therefore, coverage remains being zero.

The improvements obtained using \alg are the result of enabling better separability between query-to-relevant and query-to-irrelevant predictions. This effect can be observed in Figure~\ref{fig:distributions}, where we present a histogram of top-1 prediction scores for PRIME without and with the proposed strategy ($\mathcal{L}_{\text{base}}^{\text{TCM}}+\mathcal{L}_{xe}$). Vanilla PRIME has a strong overlap between relevant (positive) and irrelevant (negative) similarity scores, which hinders score thresholding and, therefore, its applicability to real-world systems. However, the addition of the proposed regularization decreases the negative scores, while increasing the similarities of positives. This situation in which the mode of the negatives and the main mode of the positives are greatly separated facilitates thresholding and, consequently, improves coverage. Nevertheless, some positive scores are also reduced, overlapping with the negative mode. This suggests that there is still room for improvement to find an optimal separability.

Furthermore, we present in Table~\ref{table_regs_ablation} the impact of individually adding TCM and \alg to PRIME method, demonstrating a large improvement in coverage on both datasets. 
TCM improves coverage considerably and harms precision. In contrast, \alg reduces the negative impact on precision while still boosting coverage values. When adding both regularizations together, the complementary behavior of the TCM and \alg provides a favorable trade-off between the improvement in coverage and the decrease in precision.

Overall, we have shown that the weak coverage of state-of-the-art XMC methods can be greatly improved by using our proposed auxiliary learning \alg together with the TCM regularization, a key factor for exploiting these methods in real-world systems.

% We compare the performance in terms of Precision (P@1) and Coverage (C@1) for the top item with and without regularizers on both LF-AmazonTitles-131K and Tech\&Durables datasets in Table~\ref{tabla_main_results}.

\begin{table}[t]
\begin{center}
\begin{tabular}{lcc}
\hline
{\bf Method} & {\bf P@1} & {\bf C@1} \\ \hline
\multicolumn{3}{c}{\bf LF-AmazonTitles-131K (public)} \\ \hline
\small{PRIME ($\mathcal{L}_{\text{base}}$)}  & \textbf{43.26} & 1.77\\ \hline
\small{PRIME ($\mathcal{L}_{\text{base}} \ +\ \mathcal{L}_{xe}$)} & \underline{42.14} & 9.20\\
\small{PRIME ($\mathcal{L}_{\text{base}}^{\text{TCM}}$)} & 38.70 & \textbf{14.63}\\
\small{PRIME ($\mathcal{L}_{\text{base}}^{\text{TCM}} \ +\ \mathcal{L}_{xe}$)} & 40.26 & \underline{14.59}\\ \hline
\multicolumn{3}{c}{\bf Tech\&Durables (private)}\\ \hline
\small{PRIME ($\mathcal{L}_{\text{base}}$)}  & \underline{80.83} & 46.74\\ \hline
\small{PRIME ($\mathcal{L}_{\text{base}} \ +\ \mathcal{L}_{xe}$)} & \textbf{80.92} & 54.40\\
\small{PRIME ($\mathcal{L}_{\text{base}}^{\text{TCM}}$)} & 79.40 & \underline{62.48}\\
\small{PRIME ($\mathcal{L}_{\text{base}}^{\text{TCM}} \ +\ \mathcal{L}_{xe}$)} & 80.05 & \textbf{65.08}\\ \hline
\end{tabular}
\end{center}
\caption{\label{table_regs_ablation}Performance of PRIME method when adding \alg or TCM. Both methods improve Coverage (C@1) with relatively small drops in P@1. Coverage is computed based on target P@1 of 85\% for LF-AmazonTitles-131K and 95\% for Tech\&Durables.}
\end{table}

\section{Conclusions}

This paper reveals significant limitations of product recommendation engines in real-world systems, particularly XMC methods, which exhibit low coverage rates.
To mitigate this issue, we propose \alg, a novel auxiliary learning strategy to boost the coverage of product recommendation algorithms. In particular, we regularize the training by learning to discriminate positives from the hardest in-batch negatives by means of two predictive auxiliary tasks. The first one directly uses query and label embeddings, while the second one exploits batch-aware query-label pair representations. Additionally, we demonstrate that the TCM regularization, originally proposed to palliate the threshold inconsistency problem for image retrieval, can be applied for product recommendation, and when combined with \alg, it achieves state-of-the-art results.
Our proposal to jointly regularize training via \alg and TCM strikes a balance between coverage and precision, laying the groundwork for future research in product-to-product recommendation in real-world production systems.
%drives the coverage of these systems a step forward.

\bibliographystyle{fullname}
\bibliography{EjemploARTsepln}

%\appendix
%\section{Appendix 1: TBD} TBD

\end{document}